\documentclass[a4paper]{article}

\usepackage{INTERSPEECH2021}
\usepackage{ulem}
\title{CNN Encoding of Acoustic Parameters for Prominence Detection}
\name{Kamini Sabu, Mithilesh Vaidya, Preeti Rao}
\address{
  Department of Electrical Engineering,\\
Indian Institute of Technology Bombay, Mumbai, India
  }
\email{kaminisabu@ee.iitb.ac.in,mithilesh.vaidya@iitb.ac.in,prao@ee.iitb.ac.in}

\begin{document}

\maketitle
\begin{abstract}
%Emphasis is an important factor of expressiveness used to enhance linguistic information. 
Expressive reading, considered the defining attribute of oral reading fluency, comprises the prosodic realization of phrasing and prominence.  %The automatic detection of prominence in speech has a number of practical applications. 
In the context of evaluating oral reading, it helps to establish the speaker's comprehension of the text. We consider a labeled dataset of children's reading recordings for the speaker-independent detection of prominent words using acoustic-prosodic and lexico-syntactic features. A previous well-tuned random forest ensemble predictor is replaced by an RNN sequence classifier to exploit potential context dependency across the longer utterance. Further, deep learning is applied to obtain word-level features from low-level acoustic contours of fundamental frequency, intensity and spectral shape in an end-to-end fashion. Performance comparisons are presented across the different feature types and across different feature learning architectures for prominent word prediction to draw insights wherever possible. 
    
    %The prosodic structure is important in order to understand the text and includes phrasing and prominence. This can be especially helpful in predicting the comprehension level of children as they read certain age appropriate text. Prominence is mainly used to contrast with given information or indicate new information. It is realized through sudden variations in acoustic contours. Most of the research work in prosodic event detection domain is based on use of word-level aggregated features. This work attempts to replace the handcrafting procedure by deep learning approaches.
%   The total length of the abstract is limited to 200 words. The abstract included in your paper and the one you enter during web-based submission must be identical. Avoid non-ASCII characters or symbols as they may not display correctly in the abstract book.
\end{abstract}
\noindent\textbf{Index Terms}: word prominence, children's speech, oral reading evaluation

\section{Introduction}\label{sec:intro}
The prosodic structure of speech carries important information in terms of the syntax and the meaning, both of which are critical to a listener's ease of comprehension of the spoken message~\cite{1983bock_MC_comprehension,2017maastricht_is_intonation,2017levis_pslt_infoStructure}. Phrase boundaries embed sentence syntax through word grouping while prominence or emphasis on specific words signals new information or highlights a contrast.  %While there are languages that use lexical cues such as word order to signal prominence, languages such as English employ mainly acoustic-prosodic cues. 
%The present work is motivated by the task of automatic evaluation of reading skills. 
In this paper, we investigate methods for the automatic detection of prominent words in the oral reading of middle-school children in the context of second language learning. 
The appropriate use of prosody is indicative of the reader's comprehension of the text and hence constitutes a critical component of oral reading evaluation systems~\cite{2017sinambela_ALLS_readingfluency,2017paige_JER_prosodic_reading,2019keskin_IJELS_prosodyforcomprehensibility}. %Beginning readers understand text by listening to themselves read and hence prosody aids in the comprehension [ref].  
In beginning readers, suprasegmental skills take longer to develop compared to word decoding ability, with phrasing coming earlier than the effective use of prominence. 

Prominence is perceived by a listener when a word stands out of its local context in one or more of the suprasegmental attributes such as duration, F0, intensity and spectral shape \cite{2010breen_LCP_infostructwithacoustics}. The local context itself refers to the phones and syllables within the word as well as a neighborhood of up to several words. Prosody perception is influenced not only by the low-level acoustic cues but also top-down expectations from lexico-syntactic information~\cite{2010cole_LP_prominence,2012wagner_is_perception,2018baumann_JP_prominPoSgerman}. The precise combination and relative importance of the cues depends on the speaker, language and speaking style as also on the listener. Various aggregates of the sampled acoustic parameters across the word segment including mean and variance, contour shape descriptors, and differences in these quantities across neighboring words comprise word-level prosodic features. These features are then used to train a conventional supervised classifier for the automatic detection, possibly in combination with lexico-syntactic information~\cite{2009rosenberg_thesis_prosodiceventclassification,2012mishra_is_wordprominence,2014christodoulides_avanzi_is_prominence,2015black_narayanan_is_nonnativeaccent}. In our own recent work, we applied systematic feature selection within and across the distinct suprasegmental attributes in a random forest ensemble predictor to derive a compact set of interpretable features for speaker-independent boundary and prominence detection on a children's oral reading dataset~\cite{2021sabu_CSL_prosodicevent}. With the search space for such `hand-crafted' features being very large, however, the process can miss potentially important features. Further, the pre-selected context windows used in such analyses make it difficult to exploit the long and variable time scale of prosodic relationships across an utterance in any comprehensive manner.  The potential for deep learning solutions has therefore been recognized for some time but incorporated successfully in the prominence detection task only more recently, as briefly reviewed next.

Rosenberg et al.~\cite{2015rosenberg_is_rnn} used a large number of acoustic-prosodic features and aggregates at word level derived from their previous work \cite{2009rosenberg_thesis_prosodiceventclassification,2010rosenberg_is_AuToBI} (which gave rise to the AuToBI tool) in a BiRNN classifier where the word sequence context was learned over that explicitly provided in the feature vector. They observed a small improvement ($<$ 1\% absolute) in boundary and pitch accent detection over a baseline conditional random forest classifier.
Wu et al.~\cite{2019wu_ssps_pitchaccent} also used similar aggregated acoustic features in an LSTM to find an improvement over the use of an SVM classifier. Lin et al.~\cite{2020lin_is_prosodicevents} used a hierarchical BLSTM network to aggregate features across phone, syllable and word to model contextual information at multiple granularities in the joint detection of boundaries and prominence. %Wu et al.~\cite{2019wu_ssps_pitchaccent} also evaluated the addition of lexico-syntactic features and report the following prominent word detection accuracies 65\% with acoustic features, 78\% with lexico-syntactic alone and 89\% with the combined feature set on a radio news dataset.  The higher predictive power of lexico-syntactic information such as part-of-speech tags has been noted for similar corpora before~\cite{2004chen_cole_sp_prosodylabel} [Ananthakrishnan ref].

%Motivated by the demonstrated potential of convolutional neural networks to learn discriminative patterns and replace any feature engineering in multiple speech and audio applications, recent research has focused on extracting CNN based feature representations from low-level acoustic-prosodic feature streams to obtain the word-level detection of pitch accents in an utterance [Stehwien, Zhang, Nielsen]  
In a departure from pre-computed word-level features, a recent work ~\cite{2017stehwein_is_CNNforprosodicevent,2020stehwien_SC_CNNforprosodicevents} uses CNN on frame-based acoustic parameters (energy, F0, loudness, voicing probability, zero crossing rate and harmonic-to-noise ratio) together with a context window of two neighboring words to optimally learn the high-level aggregation features. The max-pooled CNN feature maps are directly classified with a softmax layer. With word position indicators provided in the input segment, they report an improvement of 1-3\% points absolute over Rosenberg~\cite{2009rosenberg_thesis_prosodiceventclassification} on lexical stress and phrase boundary detection on the BURNC corpus, with speaker-independent scenarios being more challenging. %but report a worse performance on the BDC corpus. In general, they find within corpus training-testing results to be very high but see a relatively large reduction in cross-corpus as well as speaker independent scenarios. %A major focus of their paper is to examine what the CNN learns in the prosodic event detection task by examining the correlations of CNN output features with various hypothesized hand-crafted features.  
Zhang et al.~\cite{2018zhang_iscslp_emphasisinDialogue} also use acoustic contours and MFCC features over 10 s segments as inputs to a CNN with fixed narrow kernel width of 3 frames, with syllable and word position indicators marked at the frame level. The CNN outputs go to a BLSTM classifier to obtain emphasis at frame level. 

Given the significance of both local acoustic features and longer, more global, contexts spanning several words and possibly different sentences across the utterance in the perception of prominence, it is reasonable to consider architectures combining low-level feature aggregation with sequence models. The CNN learned feature representation of Stehwien et al.~\cite{2017stehwein_is_CNNforprosodicevent} was extended recently by Nielsen et al.~\cite{2020nielsen_emnlp_prominenceCNN} using full utterances as input and adding an LSTM layer to incorporate greater context. An improvement in accuracy of 1\% was noted for pitch accent detection on the BURNC corpus, with a further similar increase when they added text features by concatenating various dimension Glove word embeddings with the CNN embedding at the LSTM input. 

Inspired by the above reviewed works, we investigate specific enhancements to our previously proposed random forest based prominence prediction for children's oral reading evaluation, with its highly optimized acoustic-prosodic word level features~\cite{2021sabu_CSL_prosodicevent,2018sabu_sp_prosodyfeatures}. The same dataset of children's read stories in English is used with its 42,138 words across 800 utterances by 35 speakers, recorded at 16 kHz sampling and manually transcribed at word level. The selected speakers have reasonable word decoding ability in English (as second language) but widely varying levels of prosodic skill. The individual utterances comprise between 50-70 words, each word labeled for the presence/absence of prominence by 7 naive listeners using RPT~\cite{2017cole_mahrt_roy_CSL_rpt}. With a binary prominence decision based on 3 or more votes, we obtain a reasonable figure of 24\% of the total words labeled prominent. We begin with replacing the random forest predictor with an RNN sequence model that can, in principle, capture implicit context dependence from across the utterance. Next, lexico-syntactic features based on the linguistic correlates of prominence are included, with new features related to the canonical structure of the text. Finally, motivated by the demonstrated potential of convolutional neural networks to learn discriminative patterns and thus replace any feature engineering, we investigate CNN architectures for our task in the end-to-end learning of prosodic word-level features from low-level acoustic parameters.

\section{RNN-based prominence scoring}\label{sec:word-level}

Using utterances segmented via forced alignment with the transcript, Sabu and Rao~\cite{2021sabu_CSL_prosodicevent} obtained a compact and highly optimized set of word-level features by applying random forest model based feature selection on the children's speech dataset considered here. A very large set of acoustic-prosodic features defined across the different suprasegmental attributes of pitch, duration, intensity and spectral balance were considered, with multiple ways of defining temporal context in a $\pm$2 word neighborhood, to obtain a reduced set of 34 features. A Pearson correlation of 0.69 was obtained between the random forest regression output and the degree of prominence based on proportion of rater votes, and detection F-score of 0.63, in the speaker-independent prominent word detection task. In this section, we report work on using the same set of acoustic-prosodic features with an RNN model with its input given by the variable length sequence of words across an input utterance. We consider further the inclusion of lexico-syntactic information in the input sequence.
\subsection{Architecture}
%*A short section on what the below features are should be mentioned before the below line*
%We input the word level acoustic features to a sequential model similar to~\cite{2015rosenberg_is_rnn}. 
We tried different RNN architectures: GRU~\cite{ChungGCB14} and LSTM~\cite{HochSchm97} in both unidirectional and bidirectional configurations. The number of layers were varied in the range \{1, 2, 3\} while the number of hidden units were picked from the range \{48, 96, 128, 256, 512\}. At each time-step, a feature vector corresponding to a single word is fed as input to the model. A final feed-forward layer linearly transforms the RNN output at each time step to a scalar, which is passed through a sigmoid layer %\textbf{and scaled by 7} 
to get the degree of prominence prediction. %final score \textbf{in range 0-7}.
%Further, we also try adding word-level lexical and information structure related features to the 34 word-level acoustic features and then feed to the encoder architecture.

\subsection{Adding lexical features}

The high predictive power of lexico-syntactic information, such as part-of-speech tags, has been noted for several corpora before~\cite{2004chen_cole_sp_prosodylabel,2008ananthakrishnan_narayanan_ASLPTran_prosodicevent}. Baumann~\cite{2018baumann_JP_prominPoSgerman} investigated a wide range of non-prosodic factors  for prominence prediction in German speech with random forest based feature selection. %These included part-of-speech (PoS) and the number of phones and syllables in the word. %Supported by previous studies confirming the higher perceived prominence of content words, and further differences within the categories,
%Baumann~\cite{2018baumann_JP_prominPoSgerman} 
Lexical information included 12 PoS tags: NN (noun), NP (proper noun), JJ (adjective), RB (adverb), VB (verb), AU (auxiliary verb), MD (modal verb), PR (pronoun), IN (preposition), CC (conjunction), RP (particle) and DT (article). Assuming that this fine break-up may not be suited to our dataset of learners who do not necessarily have high levels of text comprehension, we also consider more coarse groupings of different dimensions. A one-hot encoding is used for the tags per word with the PoS itself determined from English Grammar rules~\cite{wren_martin_englishgrammar} since automatic parsers showed poor performance. We also include the number of phones and syllables per word which indirectly relates to word frequency~\cite{2018baumann_JP_prominPoSgerman}.   %and with the help of Oxford dictionary available on Google.
%Word level acoustic features are used for incorporating duration and pause related information. These include: pause duration before and after the word, word duration, average syllable duration and the longest syllable duration across the word. Here, pause duration is the duration of pause if it is greater than 200 ms, else it is considered as 0. These are also normalized by speech rate across the utterance. Pause duration values are also normalized by average pause value across the utterance. 

%Besides, we also use part of speech information one-hot-encoded with 12 tags similar to~\cite{2018baumann_JP_prominPoSgerman}: NN (noun), NP(proper noun), JJ (adjective), RB (adverb), VB (verb), AU (auxiliary verb), MD (modal verb), PR (pronoun), IN (preposition), CC (conjunction), RP (particle), DT (article). 
We further propose the canonical information structure (i.e. expected prominence and phrase boundaries) as additional useful features linked to top-down cues. We determine this by applying syntax and givenness rules~\cite{2017levis_pslt_infoStructure}. Motivated by the observation that expected prosodic events depend on reading speed, the events are labeled mandatory, optional and forbidden. With no known NLP methods for the automatic extraction of this information from arbitrary text, we use the model reading of the story to validate our labels.
 %Other lexical features include information structure for canonical story indicating whether a word is expected to be associated with a prosodic event for better comprehensibility. 
%For this, each story is read by an English expert at fast and slow pace. The prosodic events realized by the narrator are marked by one of the authors. 
%The information structure is encoded for each word as: the event not realized for any pace (0), the event realized for any one of the paces (1), and the event realized for both the paces (2). 
With reading miscues, albeit few, being a part of our dataset, the PoS and information structure tags are based on the target word after achieving the automatic alignment of the transcript with the text. %Since the students may not read the text correctly, there are cases of word substitutions, deletions and insertions. We align the words with canonical text and consider the most phonetically similar words as substitutions. The PoS tag and information structure for substituted words are the same as that of corresponding canonical word. For inserted words, we define separate PoS tag `i', while marking the information structure as `optional' (1). 
Other lexical features such as the number of phones and number of syllables in the word are estimated for the uttered word rather and inserted words are marked with a uniform PoS tag.

%The word frequency is known to affect the emphasis probability for a word, with commonly used words being rarely emphasized. We determine the frequency of each canonical word from a Kaggle dataset for spoken English available at~\cite{b}. The inserted words which are not available in the dataset are marked as having frequency 1. 

%Each speaker is known to have own style of speaking and use certain acoustic features more dominantly than others. We also try using the feature group importance scores from~\cite{2021sabu_CSL_prosodicevent} obtained by training the classifier on the data from a particular speaker. These are considered as the speaker embeddings. In order to get the speaker characteristics, we decide to use speaker specific characteristics like speech rate, number of pauses normalized by duration, average harmonics-to-noise ratio, pitch range, pitch variation periodicity. We also try to train a speaker classification algorithm using the 15 acoustic contours and use the bottleneck layer values as the speaker information vector.

\section{Learning word-level features with CNN} \label{sec:cnn}
\begin{figure}[t]
  \centering
  \includegraphics[width=\linewidth]{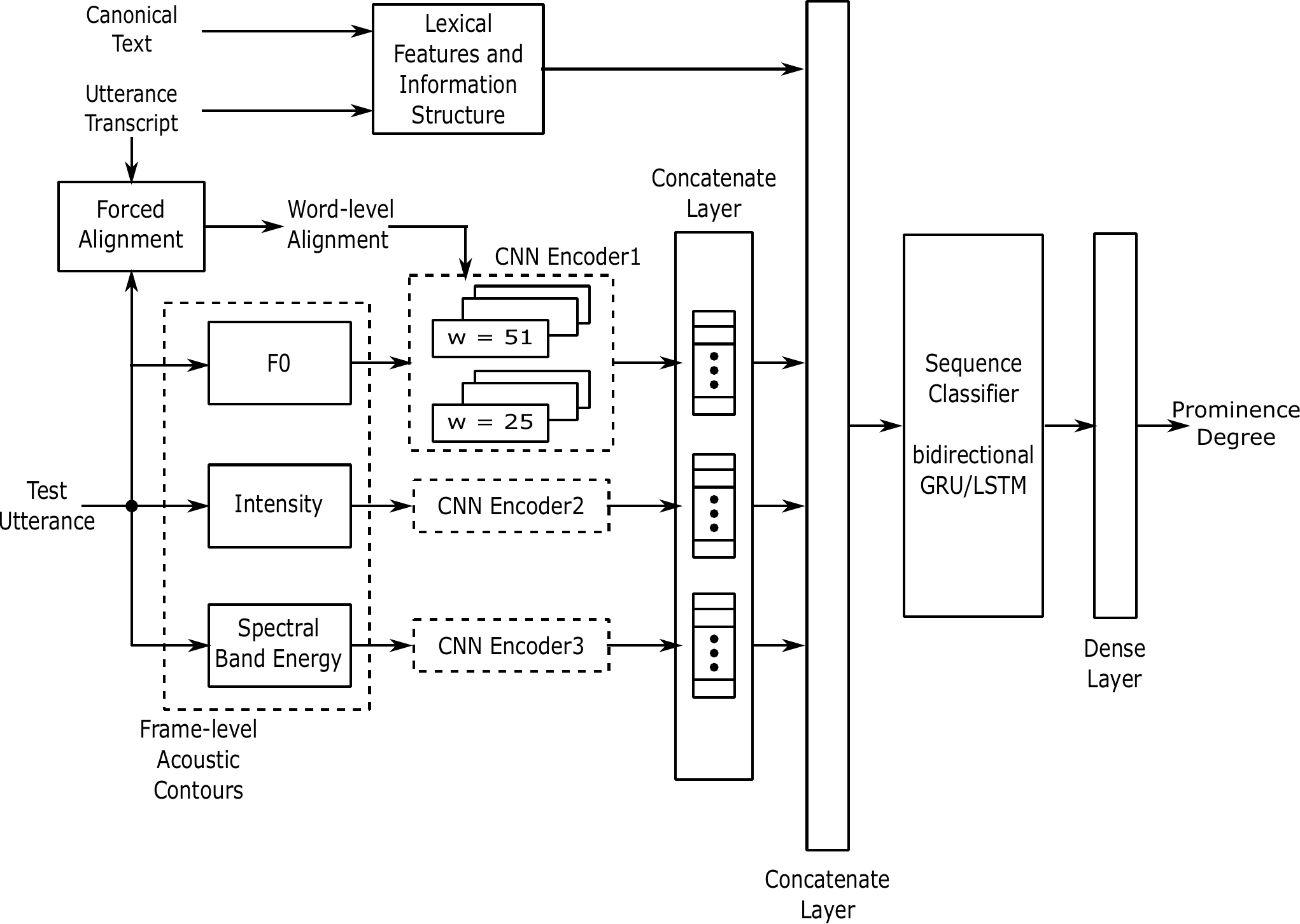}
  \caption{Prominence degree prediction architecture.}
  \label{fig:BD}
\end{figure}

%\subsection{Features}
The word-level features of our previous work were computed from word (and sub-word) aligned contours corresponding to the time-varying acoustic parameters of F0, intensity and spectral shape, computed at 10 ms intervals~\cite{2021sabu_CSL_prosodicevent}. Utterance based z-score normalization is applied on F0 and intensity parameters. We wish to investigate CNN-based automatic learning of word-level features from the same low-level acoustic contours. Given the previously observed speaker-dependence of the relative importances of the different prosodic attributes, we investigate a 3-channel CNN architecture where attribute-wise embeddings are computed with their own best filters and concatenated for the final representation~\cite{2018zhang_iscslp_emphasisinDialogue}. %as shown in Figure~\ref{fig:BD}. The specific contours in each channel are: TBD.importance, viz. pitch (F0), intensity (I), harmonic-to-noise ratio (HNR), spectral Tilt (1st MFCC), five spectral band intensities (sonorant band 300-2300Hz ($I_{sono}$), band 60-400Hz ($I_1$), band 400-2000Hz ($I_2$), band 2000-5000 Hz ($I_3$), band 5-8kHz ($I_4$)).  We considered the architecture from~\cite{2018zhang_iscslp_emphasisinDialogue}, where as in ~\cite{2018zhang_iscslp_emphasisinDialogue}. 
The contour groups are F0 (4 contours), intensity (4 contours) and spectral shape including HNR and spectral band energies (7 contours) and each feature group is input to separate CNN filter bank as shown in Figure~\ref{fig:BD}.
The output encodings from the filter banks are finally concatenated and fed to the sequence classifier (GRU/LSTM) network together with other word-level features as considered in our experiments.

 Perceived prominence depends on the underlying temporal dynamics while CNN computations are time-invariant. To match the information available to the hand-crafted feature extraction, the CNN input is a segment with the selected word context, possibly with word and subword positions within the segment provided explicitly. We explore a range of context choices as also different types of position information, starting with the context of $\pm$ 1 word given the importance of immediately neighboring words in the realization of emphasis on a word~\cite{2021sabu_CSL_prosodicevent,2020stehwien_SC_CNNforprosodicevents}.
%For ZCR, F0 and I, the contours are considered in both raw and z-normalized form. %For pitch, semitones with respect to the average pitch across recording are also considered. Of these, ZCR was observed to reduce the performance. Therefore, we finally considered only 15 contours for the task.
%\subsection{Architecture}
%We use a multi-channel CNN, similar to Zhang~\cite{2018zhang_iscslp_emphasisinDialogue}, with each of the 3 channels corresponding to the acoustic contours of F0, intensity and spectral band energies. We however use variable duration segments as input to the CNN. 
%Multiple filter kernel sizes are used giving rise to hyperparameters that need to be tuned on training data. The architecture used for this work is similar to \textbf{(?)}. The acoustic contours are fed to a Convolutional Neural Network (CNN), one word at a time. 
We also consider CNN filter banks with a range of kernel width choices motivated by sub-word units~\cite{2018trang_ostendorf_nnaclhlt_scoring}. %The initial set of 4 kernel widths are: 5 frames corresponding to a sub-phone context, 11 frames covering a phone, 31 frames for a syllable and 51 frames for a word. 
The 1D convolution output of each of the CNN filters is max-pooled across time to get a scalar value per filter per filter bank. Each filter bank has N filters, each with k kernel sizes, resulting in a kN-dimensional feature encoding, for each channel, corresponding to a word.% as shown in Figure~\ref{fig:BD}. 
%This word level encoding is then concatenated with word level acoustic features related to duration and pause which were not considered during the CNN encoding of the architecture.
%Further, lexical features are also added to the word-level feature vector and the vector is input to the sequential classifier (BLSTM/BGRU encoder).
%\textbf{(We have mentioned 4N because we were using 4 widths. Should we replace it with kN where k is the number of filter widths? e.g.w e are using only 2 later)YES}

%In order to know if new data helps, we use the unlabeled data in teacher-student configuration. We predict the ratings for this data using our RF model from~\cite{2021sabu_CSL_prosodicevent}. These labels are then used to further train the model trained using the labeled data.

\section{Experiments and results}\label{sec:results}

\subsection{Train-test splits and baseline}
%We use an available dataset ~\cite{2021sabu_CSL_prosodicevent} of children's read stories in English with its 41,326 words by 35 speakers, recorded at 16 kHz sampling. The individual utterances comprise between 50-70 words, each labeled for the presence/absence of prominence by 7 raters using RPT [Cole].
%The dataset comes from . The students have Marathi as native language and are in English. The reading data are story texts, split into paragraphs of about 50–70 words, presented in plain Roman script with the normally expected punctuation. Student reads aloud one paragraph at a time with a headset microphone. Recordings are stored at 16 kHz sampling frequency in 16-bit PCM format. The recording sessions were conducted in relatively quiet environment.

%Dataset comes from recordings of stories read by 10-14 year old students from different schools in Maharashtra~\cite{2021sabu_CSL_prosodicevent}. There are 921 recordings from across 35 students transcribed and annotated for presence of prosodic event by multiple raters of which we consider the data rated by 7 raters. Each word is assigned a ground-truth label between 0 and 7 for each of the prosodic events depending on the number of votes it obtained for event presence. We have 41,326 words from 791 recordings by 35 speakers for phrase boundary detection and 42,138 words from 808 recordings by 35 speakers for the prominent word detection task. Inter-rater agreement among the 7 raters measured in Fleiss' kappa is 0.59 and 0.22 for phrase boundary and prominence marking respectively.

The random forest predictor based system reported in~\cite{2021sabu_CSL_prosodicevent} serves as a baseline and we adopt the identical training and testing methodology here. The complete dataset of 42,138 words is split into three equal folds with no speaker overlap for 3-fold cross-validation based testing. The hyperparameters are tuned with 4-fold CV on the train split. The model is then trained on the entire train set and results are reported on the corresponding unseen test set. We report the mean and standard deviation across the three test folds. The results for the prominence degree prediction are reported in terms of Pearson correlation between the predictor output and the degree of prominence from the RPT rater votes. Also, the prominence detection F-score values are reported considering prominence present when a word receives 3 or more votes as discussed in Section~\ref{sec:intro}. %where the regression output and the ground truth votes are thresholded at 2 or 3 votes.

\subsection{RNN training and performance}

%The baseline model comes from~\cite{2021sabu_CSL_prosodicevent}, where a RF regressor is used to predict the degree of prominence using an optimal set of carefully selected 34 handcrafted features. 
We present the entire input utterance in the form of a sequence of word-level features to an RNN classifier. Various feature combinations are investigated to determine the individual and combined contributions of prosodic and lexical features. We test various RNN architectures as presented in Section~\ref{sec:word-level}. %GRU~\cite{ChungGCB14} and LSTM~\cite{HochSchm97} in both unidirectional and bidirectional configurations. The number of layers were varied in the range {1,2,3} while the number of hidden units were picked from the range {48, 96, 128, 256, 512}.
%\textbf{(This is repeated from 2.1). Can skip this range of hyperparameters} 
%try replacing the RF model with other sequential model similar to~\cite{2015rosenberg_is_rnn}. 
For training, we used the AdamW~\cite{adamW} optimizer, an improved version of popular Adam~\cite{kingma2017adam} optimizer with a weight decay mechanism that helps with faster training and more generalized models. We used a learning rate of 0.003 and a batch size of 500. Dropout with probability 0.2 is added to each RNN layer except the last layer. The Mean Squared Error between the scaled target score (between 0 and 1) and the predicted score is used as the loss function to be minimized during training. 
% AdamW estimates lr using first and second order moments

\begin{table}[th]
\footnotesize 
  \caption{Performance of various models with set of 34 acoustic features. (* indicates sd $<$ 0.01)}
  \label{tab:RFvsRNN}
  \centering
%   \begin{tabular}{ r@{}l  r }
    \begin{tabular}{l|l|l|l|l}
    \toprule
    Model & \# layers & \# units & Correlation & F-score \\
    % \multicolumn{2}{c}{\textbf{CNN context}} &                                  \multicolumn{1}{c}{\textbf{CNN filters}} \\
    \midrule
    RFC & - & - & 0.69* & 0.63* \\
    GRU & 2 & 96 & 0.68~~ & 0.63~~ \\
    LSTM & 2 & 256 & 0.69~~ & 0.63~~ \\
    BGRU & 2 & 96 & 0.70~~ & 0.64~~ \\
    BLSTM & 2 & 256 & 0.71* & 0.64* \\
    \bottomrule
  \end{tabular}
\end{table}
Table~\ref{tab:RFvsRNN} shows the performance with the 34 acoustic-prosodic features (`A34') used in the RNN architectures compared with the same features in the baseline random forest predictor. We note that there is an improvement from the utterance-long context available to the sequence models, 
especially in the case of the bidirectional models. We employ the BGRU going ahead due to it ease of training and known suitability for lower dataset sizes. %Here, the GRU model is 2-layer bidirectional GRU with 96 hidden units \textbf{(No need to mention; given in the table?)}. 

\begin{figure}[t]
  \centering
  \includegraphics[width=0.5\linewidth]{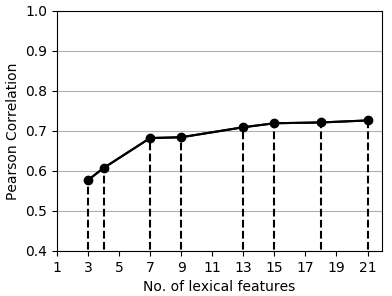}
  \caption{Correlation performance variation with different sets of lexical features.}
  \label{fig:lexical}
\end{figure}

Next, considering separately the lexical features, Figure~\ref{fig:lexical} shows performance for various reduced sets starting from the full set of 21 that includes PoS tags and phone/syllable counts (together termed `L'), and information structure labels (termed `I'). From a maximum correlation of 0.72, we note a drop in performance as the 6 information structure and 2 word length features are removed to get to 13 features corresponding to the PoS alone. A further drop is recorded when the PoS tags are grouped in different ways to get to the final feature set of only content/function word distinctions. We note that lexical features alone show a predictive power similar to prosodic features alone (and not much higher as in some previous work~\cite{2019wu_ssps_pitchaccent,2008ananthakrishnan_narayanan_ASLPTran_prosodicevent}), explained by the lower proficiency speakers of our dataset. All the same, the reduction of PoS tags clearly hurts performance. Table~\ref{fig:lexical} summarizes the achieved performance gains as we augment the acoustic-prosodic feature set with lexical (all but information structure) and information structure features with each of the latter two clearly adding value.
%\textbf{where is only L? table contains A34+L} with  %As we tried combining some of the attributes thus reducing the details in the information structure, we get  a trend as shown in 

%Further, concatenating the lexical and acoustic features, the performance improves even further indicating that the two contain complementary information. 

\begin{table}[th]
\footnotesize
  \caption{Performance with addition of lexical and information structure features. (* indicates sd $<$ 0.01)}
  \label{tab:lexicalRNN}
  \centering
%   \begin{tabular}{ r@{}l  r }
    \begin{tabular}{l|l|l}
    \toprule
    % feature & Pearson & F-score & Pearson & F-score \\
    % group & corr & ($\geq3$) & corr & ($\geq3$) \\
    
% A34 & 0.71 & 0.64* & 0.68 & 0.63 \\
% % A34+ I & 0.76* & 0.67* & 0.54 & 0.58 \\
% A34+ L & 0.78* & 0.68* & 0.78* & 0.69* \\
% A34 + L + I & 0.80* & 0.70* & 0.79* & 0.69* \\
Features & Correlation & F-score \\
\toprule
% A24 & 0.68* & 0.62* \\
A34 & 0.70~~ & 0.64~~ \\
% A24 + L + I & 0.78* & 0.69* \\
A34 + L & 0.75* & 0.67* \\
A34 + L + I & 0.79* & 0.69* \\
    \bottomrule
  \end{tabular}
\end{table}

%In order to replace the tedious handcrafting procedure with deep learning technique, we use CNN similar to \cite{2018trang_ostendorf_nnaclhlt_scoring,2018zhang_iscslp_emphasisinDialogue,2020stehwien_SC_CNNforprosodicevents}. 

\subsection{CNN training and performance}

While our multi-channel CNN framework is similar to that of Zhang et al.~\cite{2018zhang_iscslp_emphasisinDialogue}, we expand the search for architecture choices by considering the use of multiple kernel widths in each channel to capture the distinct time scales of acoustic variation.   %involved use of single filter kernel width with size 3 of 3 frames. 
%We expect that use of multiple filter widths corresponding to different acoustic sub-word units may lead to performance improvement. 
We start from the 4 kernels with widths [5, 11, 25, 51] similar to that of the sentence parsing CNN architecture of Trang et al.~\cite{2018trang_ostendorf_nnaclhlt_scoring}, which roughly cover sub-phone, phone, syllable and word, and possibly some context. Given the fixed narrow kernel width of 3 frames used in~\cite{2017stehwein_is_CNNforprosodicevent,2018zhang_iscslp_emphasisinDialogue}, we add this to our candidates for testing. From the different combinations presented in Table~\ref{tab:kernelsize}, we observe that the syllable and word width kernel sizes (25, 51) helps the performance while including other widths does not change it. For these instances, we fixed the number of filters of each kernel size to 8. The input context to CNN is also fixed to $\pm$ 1 word with position encoding used to indicate current word based on~\cite{2020stehwien_SC_CNNforprosodicevents}. We find that phone and sub-phone width kernels do not help, and even degrade the performance in some cases.
We finalize the use of two kernel widths 25 frames and 51 frames, corresponding to syllable and word widths, for each CNN filter bank. Next, to find the optimal number of kernels for each width, we varied the number of CNN filters in the range \{4, 8, 16, 20\}. We found that 8 filters give the best performance and adopted this for all the further experiments.

\begin{table}[th]
\footnotesize
  \caption{Performance for different sets of kernel width choices with the corresponding lit. reference indicated.} 
  %We use 8 filters in each CNN filter bank. Input to CNN has $\pm$ 1 word context with position encoding to indicate current word similar to~\cite{2020stehwien_SC_CNNforprosodicevents}.
%   (* indicates s.d. $<$ 0.01)}
  \label{tab:kernelsize}
  \centering
%   \begin{tabular}{ r@{}l  r }
    \begin{tabular}{l|l|l}
    \toprule
    Kernel widths & Correlation & F-score \\
    % \multicolumn{2}{c}{\textbf{CNN context}} &                                  \multicolumn{1}{c}{\textbf{CNN filters}} \\
    \midrule
    3,5,11,25,51 & 0.67~~ & 0.62~~ \\
    5,11,25,51~\cite{2018trang_ostendorf_nnaclhlt_scoring} & 0.67~~ & 0.62~~ \\
    25,51 & 0.67* & 0.62* \\
    11,25,51 & 0.67* & 0.62~~ \\
    11~\cite{2020nielsen_emnlp_prominenceCNN} & 0.65~~ & 0.61* \\
    % 25 & 0.69* +- 0.0086 & 0.69 +- 0.0189 \\
    % 51 & 0.68 +- 0.0166 & 0.69 +- 0.0178 \\
    \bottomrule
  \end{tabular}
\end{table}
%Based on the results presented in Table~\ref{tab:kernelsize} while limiting the trainable parameters, 

% \begin{table}[th]
%   \caption{Pearson correlation with varying number of filters for kernel width = [25,51]. Input to CNN has $\pm$ 1 word context with position encoding to indicate current word similar to~\cite{2020stehwien_SC_CNNforprosodicevents}.}
%   \label{tab:nofilters}
%   \centering
% %   \begin{tabular}{ r@{}l  r }
%     \begin{tabular}{c|c|c}
%     \toprule
%     no. of filters & GRU & LSTM \\
%     % \multicolumn{2}{c}{\textbf{CNN context}} &                                  \multicolumn{1}{c}{\textbf{CNN filters}} \\
%     \midrule
%     4 & 0.68 +- 0.0177 & 0.68 +- 0.0179 \\
%     8 & 0.69 +- 0.0162 & 0.69 +- 0.0147 \\
%     16 & 0.68 +- 0.0101 & 0.67 +- 0.01 \\
%     20 & 0.68 +- 0.0151 & 0.68 +- 0.0137 \\
%     \bottomrule
%   \end{tabular}
% \end{table}

Next, we consider different choices for position encoding. Stehwien et al.~\cite{2017stehwein_is_CNNforprosodicevent} found it useful to indicate the frames corresponding to the word as distinct from the context frames from the neighboring words within the input segment. However, it has been observed that the preceding and following words influence perceived prominence in different ways~\cite{2021sabu_CSL_prosodicevent}. Therefore, we decided to change the position encoding to distinguish all the three words. To test whether a special encoding to indicate inter-word pauses can help, we applied 5-bit one-hot encoding across the segment of 3 words and 2 pauses. The results are shown in Table~\ref{tab:pos_enc}. Taking position information to intra-word level, we explore one-hot encoding to indicate syllable number (up to a maximum of 7 syllables) in the word as indicated in last row of the table. As can be seen, the pause encoding improves the performance thereby indicating that it is important indeed to differentiate between next and previous word as well as pauses. %\textbf{(the table has the same 2nd decimal place so cant differentiate)}. 
Syllable marking helps too, and further experiments are reported using the 5-bit encoding along with the 7-bit syllable position.

\begin{table}[th]
\footnotesize
  \caption{Performance with various positional encoding approaches. Input to CNN has $\pm$ 1 word context.}
  \label{tab:pos_enc}
  \centering
%   \begin{tabular}{ r@{}l  r }
    \begin{tabular}{l|l|l}
    \toprule
    Positional encoding & Correlation & F-score \\
    % \multicolumn{2}{c}{\textbf{CNN context}} &                                  \multicolumn{1}{c}{\textbf{CNN filters}} \\
    \midrule
    1-bit~\cite{2020stehwien_SC_CNNforprosodicevents} & 0.67* & 0.62*  \\
    3-bit (word order) & 0.68~~ & 0.62~~ \\
    5-bit (word/pause order) & 0.68~~ & 0.62~~ \\
    5-bit + syllable position & 0.68~~ & 0.63~~ \\
    % 3-bit + syllable position &  & 0.68 & 0.63 \\
    \bottomrule
  \end{tabular}
\end{table}

%With the $\pm$ 1 word context used for the task, and the limitation of CNN taking fixed size input, we need to perform zero-padding. Since the presence of silence is important cue to prominence, the zero padding may affect the results. We hope that the positional encoding can take care of it, but still it may be worth a try to chunk the recording into fixed length segments centered at the word and input these to the CNN. 
We also experimented with choosing fixed duration input segments, centered at the word, as an alternative to the variable (3 word) segments considered so far to avoid the zero-padding at the CNN input. We find that the duration that works best corresponds to the average 3-word segment and that there is no performance gain with fixing input segment length. %that For example, a segment length of 91 frames corresponds to the word and surrounding context that depends on the different word durations. Of the lengths tried  f45 equivalent to $\pm$45 frames around the word centre gives the best results.% as shown in Table~\ref{tab:context}. 
%The information available for feature learning is now dependent on the word length rather than fixed to the earlier $\pm$ 1 word.
%The maximum word length in our dataset is 800 ms. The window of 91 frames is, therefore, able to cover the current word and also consider some context around it. In case of shorter current word, the window can have more surrounding words as well and can benefit from more context. On the other hand, for longer current word, it can benefit by giving more focus to the syllables within the word. With increase in context window, however, the performance starts reducing instead.

% \begin{table}[th]
%   \caption{Pearson correlation with varying CNN input context.}
%   \label{tab:context}
%   \centering
% %   \begin{tabular}{ r@{}l  r }
%     \begin{tabular}{c|c|c}
%     \toprule
%     CNN & Correlation & F-score \\
%     % \multicolumn{2}{c}{\textbf{CNN context}} &                                  \multicolumn{1}{c}{\textbf{CNN filters}} \\
%     \midrule
%     w1 & 0.68 & 0.63 \\
%     f35 & 0.69* & 0.63 \\
%     f45 & 0.69 & 0.63 \\
%     f55 & 0.68 & 0.62 \\
%     \bottomrule
%   \end{tabular}
% \end{table}

To capture the contribution of multi-channel processing of the attribute-wise contour, we compare it with single-channel processing of the combined contours in Table~\ref{tab:arch} with all else kept unchanged. We note a drop in performance. Feature pooling of multi-channel outputs also reduced performance compared to simple feature concatenation suggesting the importance of retaining all attribute variations in the input to the RNN model. %The inter-dependence between channels is not need further investigation.

%With the optimal configuration so far, we tried two further changes in the CNN architecture. First of all, we combined all the features into a single group and used a single CNN filterbank with kernel sizes [25,51]. This would help get the inter-dependence of the acoustic contours from different groups. However, as pointed in~\cite{2018zhang_iscslp_emphasisinDialogue}, multi-channel approach turned out to be more efficient. Further, we also tried combining the three feature groups by max-pooling across the three filterbank outputs instead of concatenating. This would help the most dominant of the feature groups to take over at any time instance. However, this too led to reduced performance. This means, the individual feature groups should be dealt separately. 

\begin{table}[th]
\footnotesize
  \caption{Performance with different CNN architectures}
  \label{tab:arch}
  \centering
%   \begin{tabular}{ r@{}l  r }
    \begin{tabular}{l|l|l}
    \toprule
    Architecture & Correlation & F-score \\
    % \multicolumn{2}{c}{\textbf{CNN context}} &                                  \multicolumn{1}{c}{\textbf{CNN filters}} \\
    \midrule
    Multichannel (concatenate) & 0.69~~ & 0.63~~ \\
    Multichannel (pooling) & 0.66~~ & 0.62~~ \\
    Single-channel & 0.67* & 0.62~~ \\
    \bottomrule
  \end{tabular}
\end{table}

% \begin{table}[th]
%   \caption{Results with CNN followed by encoder for different CNN configurations}
%   \label{tab:CNN}
%   \centering
% %   \begin{tabular}{ r@{}l  r }
%     \begin{tabular}{c|c|c|c|c|c}
%     \toprule
%     CNN & pos & Pearson & F-score & Pearson & F-score \\
%     context & enc & corr & ($\geq3$) & corr & ($\geq3$) \\
%     % \multicolumn{2}{c}{\textbf{CNN context}} &                                  \multicolumn{1}{c}{\textbf{CNN filters}} \\
%     \midrule
%     w1 & & 0.27 & 0.45 & 0.24 & 0.45 \\
%     f30 & no & 0.59* & 0.58* & 0.61* & 0.59* \\
%     v20 & no & 0.65* & 0.61 & 0.63 & 0.6 \\
%     w1 & yes & 0.5 & 0.55 & 0.68 & 0.62* \\
%     f30 & yes & 0.68 & 0.62 & 0.68 & 0.62 \\
%     v20 & yes & 0.69* & 0.63* & 0.67 & 0.62 \\
    
%     \bottomrule
%   \end{tabular}
% \end{table}

% The results show that having word context w1 (current word $\pm$ 1 word context) as mentioned in~\cite{2020stehwien_SC_CNNforprosodicevents} gives poorer results as compared to the other two context mechanisms. The variable context v20 (word $\pm$ 20 frames) proves to be the best context when no positional encoding is applied. When we use 3-bit positional encoding to indicate the previous word, current word and next word, the performance improves for all the context types. Moreover, the fixed context f30 (word center $\pm$ 30 frames) proves to be the best context in this case. This indicates that in the absence of word explicit position information, variable context tried to benefit from word being at center. However, if explicit position information is provided, this is unnecessary.

\subsection{Overall performance}
Given our overall goal of investigating the automatic learning of features for a prosodic prominence detection task on a challenging dataset of children's read speech, we explored the cascade of CNN and RNN with word and sentence level inputs respectively. Our baseline was a random forest ensemble predictor with hand-crafted acoustic-prosodic features optimized with  to exploit the best of acoustic parameter aggregation, normalization and local context for prominence detection of a word within a long utterance. Using a bidirectional GRU with the sequence of acoustic-prosodic features (`A34') helped improve F-score by 1\% over the baseline. As we see in row 2 of Table~\ref{tab:CNNandword}, the CNN-learned features in the same setting fall slightly short. 

Adding features assumed to be harder to learn automatically, but important for word prominence, such as the actual and speech-rate normalized word, syllable and pause durations, we concatenate a 12-dimensional duration feature vector (`DP-12') with the CNN-learned features.  We further select 10 contour shape features (`A10', a subset of A34) that require fitting the temporal variation of the acoustic parameters. The added features are seen to bring distinct additional value. Finally, the inclusion of text features (L, I) boosts performance for both feature sets as we see in the last 2 rows of Table~\ref{tab:CNNandword} with higher performance for A34. 

Our outcomes are overall consistent with those of the few previous works that have compared automatically learned and hand-crafted features on the same dataset and task. As is well known, automatic feature learning is limited by dataset sizes and future work must examine the use of larger, possibly unlabeled, data and architectures that learn additional useful information such as speaker identity or employ attention for better context learning.
%We can also see from the results that the performance with CNN encodings is very similar to that of the 34 features given the same encoder configuration. The handcrafted features can be replaced by the CNN encoder architecture.
%As CNN is not explicitly given the duration and pause information, we concatenate 12 different duration and pause (D-P12) features with the CNN encoding. Duration features include the absolute word duration, longest and average syllable duration in the word. These are also normalized by speech rate. The pause features include the duration of the previous and successive pause normalized by speech rate and average pause duration. Further, we also add 10 other duration and contour slope features (A10) subset of the 34 acoustic features. We feel that these are not simple aggregates and hard to learn for the CNN. Both of them when added together, tend to give improvement surpassing the GRU model trained on handcrafted 34 acoustic features.

\begin{table}[th]
\footnotesize
  \caption{Performance of CNN encoding concatenated with different word-level features as RNN input. (* indicates sd $<$ 0.01)}
  \label{tab:CNNandword}
  \centering
%   \begin{tabular}{ r@{}l  r }
    \begin{tabular}{l|l|l}
    \toprule
    Features & Correlation & F-score \\
    % \multicolumn{2}{c}{\textbf{CNN context}} &                                  \multicolumn{1}{c}{\textbf{CNN filters}} \\
    \midrule
A34 & 0.70~~ & 0.64~~ \\
CNN & 0.69~~ & 0.63~~ \\
% CNN + L + I & 0.77* & 0.67* \\ %& 0.78* & 0.68* \\
% CNN + D-P12 & 0.69 & 0.63  \\
% CNN + A10 & 0.69* & 0.63 \\
CNN + D-P12 + A10 & 0.71~~ & 0.64~~ \\ %0.68* & 0.63 \\
% CNN + A10 + L + I & 0.77* & 0.68 \\
CNN + D-P12 + A10 + L + I & 0.77* & 0.68~~ \\
% CNN + durpause(12) + A34 + L + I & 0.78* & 0.69* \\
A34 + L + I & 0.79* & 0.69* \\
    \bottomrule
  \end{tabular}
\end{table}

\bibliographystyle{IEEEtran}

\bibliography{mybib}

% \begin{thebibliography}{9}
% \bibitem[1]{Davis80-COP}
%   S.\ B.\ Davis and P.\ Mermelstein,
%   ``Comparison of parametric representation for monosyllabic word recognition in continuously spoken sentences,''
%   \textit{IEEE Transactions on Acoustics, Speech and Signal Processing}, vol.~28, no.~4, pp.~357--366, 1980.
% \bibitem[2]{Rabiner89-ATO}
%   L.\ R.\ Rabiner,
%   ``A tutorial on hidden Markov models and selected applications in speech recognition,''
%   \textit{Proceedings of the IEEE}, vol.~77, no.~2, pp.~257-286, 1989.
% \bibitem[3]{Hastie09-TEO}
%   T.\ Hastie, R.\ Tibshirani, and J.\ Friedman,
%   \textit{The Elements of Statistical Learning -- Data Mining, Inference, and Prediction}.
%   New York: Springer, 2009.
% \bibitem[4]{YourName17-XXX}
%   F.\ Lastname1, F.\ Lastname2, and F.\ Lastname3,
%   ``Title of your INTERSPEECH 2021 publication,''
%   in \textit{Interspeech 2021 -- 20\textsuperscript{th} Annual Conference of the International Speech Communication Association, September 15-19, Graz, Austria, Proceedings, Proceedings}, 2020, pp.~100--104.
% \end{thebibliography}

\end{document}